\pdfoutput=1
\documentclass[11pt]{article}

\usepackage{ACL2023}

\usepackage{times}
\usepackage{latexsym}

\usepackage[T1]{fontenc}
\usepackage[utf8]{inputenc}

\usepackage{microtype}

\usepackage{inconsolata}

\usepackage{blindtext}
\usepackage{hyperref}
\usepackage{graphicx}
\usepackage[textsize=tiny]{todonotes}
\usepackage{booktabs}
\usepackage{makecell}
\usepackage{amsmath}
\usepackage{footnote}

\title{Honey, I Shrunk the Language: Language Model Behavior\\ at Reduced Scale}

\author{Vijeta Deshpande\textsuperscript{1}, Dan Pechi\textsuperscript{2}, Shree Thatte\textsuperscript{1}, Vladislav Lialin\textsuperscript{1}, Anna Rumshisky\textsuperscript{1,3} \\
  \textsuperscript{1}University of Massachusetts Lowell, Computer Science Department \\
  \textsuperscript{2}New York University, Center for Data Science \\
  \textsuperscript{3}Amazon Alexa AI \\
  \texttt{\{vijeta\_deshpande,shree\_thatte\}@student.uml.edu}, 
  \texttt{danpechi@nyu.edu} \\
  \texttt{\{vlialin,arum\}@cs.uml.edu} \\
}

\begin{document}
\maketitle
\begin{abstract}

In recent years, language models have drastically grown in size, and the abilities of these models have been shown to improve with scale. The majority of recent scaling laws studies focused on high-compute high-parameter count settings, leaving the question of when these abilities begin to emerge largely unanswered. 
In this paper, we investigate whether the effects of pre-training can be observed when the problem size is reduced, modeling a smaller, reduced-vocabulary language. 
%We examine downscaling effects for masked language modeling and 
%We pre-train ~70 models in the range of 1-100M parameters using a masked language modeling objective. 
%
We show the benefits of pre-training with masked language modeling (MLM) objective in models as small as 1.25M parameters, and establish a strong correlation between pre-training perplexity and downstream performance (GLUE benchmark). We examine downscaling effects, extending scaling laws to models as small as ~1M parameters. At this scale,  we observe a break of the power law for compute-optimal models and show that the MLM loss does not scale smoothly with compute-cost (FLOPs) below $2.2 \times 10^{15}$ FLOPs. 
%
%Interestingly, for compute-optimal models, pure parameter count is not predictive of perplexity in this setting and certain model configurations remain compute-optimal for a long range of FLOPs. 
%
We also find that adding layers does not always benefit downstream performance.\footnote{Our filtered pre-training data, reduced English vocabulary, and code are available at \href{https://github.com/text-machine-lab/mini_bert}{https://github.com/text-machine-lab/mini\_bert}.}

\end{abstract}

\begin{figure}[ht!]
\centering
\includegraphics[width=75mm]{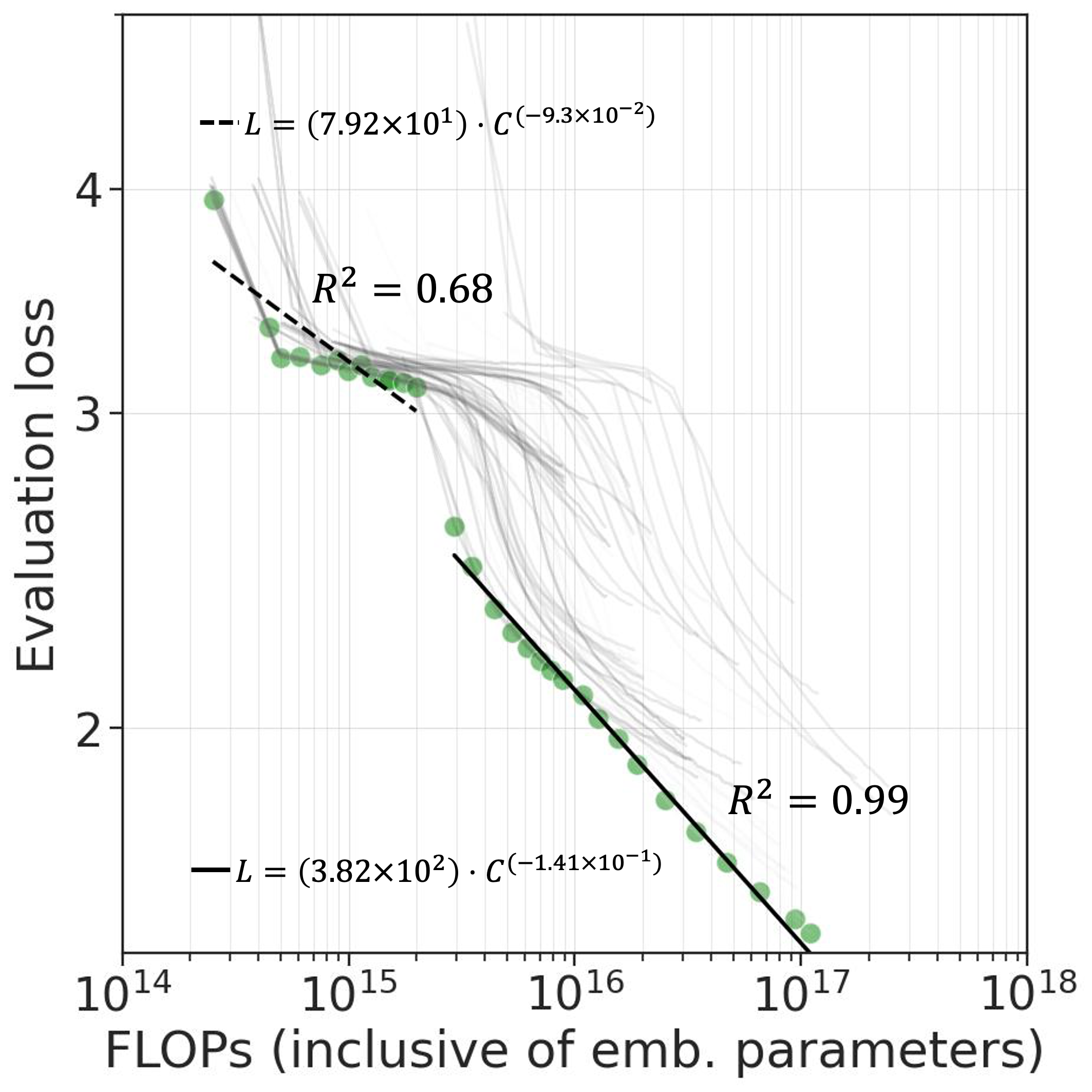}
% \caption{\textbf{Discontinuity in the power law.} In this figure we plot compute-cost (FLOPs) (on the x-axis) and corresponding MLM loss values (on the y-axis) recorded during pre-training. The faded gray lines represent unique model configurations. The green dots represent compute-optimal instances i.e., the model instance with a minimum value of MLM loss for a given range of FLOPs. The solid and the dotted black lines represent a power curve fit on the compute optimal instances. We observe that there exists a discontinuity in the power curve around the FLOPs value of $2.2 \times 10^{15}$}
\caption{\textbf{Power law breaks in low FLOPs region.} Faded gray lines are unique model configurations. The green dots are compute-optimal instances i.e., the model instance with a minimum value of MLM loss for a given range of FLOPs. We observe that there exists a discontinuity in the power curve around $2.2 \times 10^{15}$ FLOPs}
\label{figure:flops-vs-PPL}
\end{figure}

\section{Introduction}

In the past few years, large language models (LLMs) have grown ever larger \cite{brown2020_language_gpt3,shoeybi2019megatron,palm,switch_transformers}, and the emergent abilities of these models improve with scale. While several studies have looked at the relationship between model size, the amount of training, and performance for LLMs \cite{kaplan2020scaling,hoffmann-optimal}, the main focus has been on scaling laws for high-compute settings. Very few studies have considered the effects of pre-training at a smaller scale \cite{turc-students,huebner2021babyberta}. Thus, the question of when exactly model abilities begin to emerge remains largely unanswered.

In this study, we were interested in understanding whether the emergent phenomena can be observed at a drastically reduced scale, and what the relationship is between upstream and downstream performance at this scale. We also wanted to examine model shapes, configurations, and other factors that might affect whether we see the benefits of pre-training when downscaling a model.

Smaller models have been shown to do poorly when trained even on large volumes of data \cite{turc-students}, which makes studying downscaling non-trivial. However, during language acquisition, humans are exposed to a reduced-size  language before gradually expanding their vocabulary, yet they become fluent even when their vocabulary is limited. Taking our cue from humans, we explore the hypothesis that reducing language size might allow us to observe the effects of pre-training in small models.

There has been one previous attempt to reduce language size  \cite{huebner2021babyberta}, but it was quite limited: one reduced-size Transformer encoder was trained with a non-standard version of masked language modeling (MLM) loss on a relatively small corpus of child-directed speech and evaluated for its ability to pass linguistic tests from a custom grammar test suite.

% DESCRIBE WHAT WE DO BRIEFLY
We use a vocabulary of 21,000 words derived from AO-CHILDES \cite{huebner2021using}, a corpus of child-directed speech, to create a filtered corpus containing a subset of standard pre-training corpora: C4 \cite{raffel2020exploring}, Wikipedia, Book Corpus \cite{Zhu_2015_ICCV}, and others. We pre-train over 70 Transformer encoder models in the 1-100M parameter range, varying model shape, and configuration, including the number of layers, hidden size, the number of attention heads, and the feed-forward layer dimension (intermediate size). We fine-tune and evaluate a series of checkpoints at different FLOPs count on the subset of GLUE filtered with the same vocabulary.

% WHAT WE FOUND
We present evidence that for a realistically downscaled language, the benefits of pre-training are observable even in smaller models with as few as 1.25M parameters.
Our results also indicate that models with fewer layers achieve better performance on GLUE tasks. In contrast to \citet{tay2022scale}, we find a strong correlation between upstream and downstream performance (here, model perplexity and GLUE score). However, pre-training compute optimality does not appear to be crucial for downstream results. We also show that for compute-optimal models at this scale, parameter count does not reliably predict MLM loss, suggesting a limitation to scaling laws.  We observe a departure from the FLOPs-Perplexity law, characterized by a sudden shift in the exponent value in the low-compute region of FLOPs $\leq 2.2 \times 10^{15}$ (cf. Figure \ref{figure:flops-vs-PPL}). This represents a divergence from previous observations regarding scaling laws \cite{kaplan2020scaling, hoffmann-optimal}.

\section{Related Work}

\paragraph{Scaling Laws}

\citet{kaplan2020scaling} has demonstrated and popularized power law dependency between language model parameter count and perplexity. This further motivated the existing trend for increasing model size \cite{brown2020_language_gpt3,palm,switch_transformers}.
Investigation of smaller models has mostly focused on distillation \cite{sanh-distilbert,turc-students} and achieving the best performance given the parameter count. In contrast, we are interested in understanding at what scale pre-training works and the emergence of language model abilities to help downstream tasks.

Changing pre-training data size via training token volume \cite{perez-pretraining,zhang-billions}, vocabulary size \cite{gowda-may-2020-finding}, or the number of epochs \cite{Voloshina_2022} has also been explored for effects on language acquisition. These studies have generally demonstrated low-level linguistic tasks involving syntax require relatively little data volume compared to more complex tasks like WinoGrande \cite{Sakaguchi2019WINOGRANDEAA}.
The relationship between model size, input data, and downstream performance remains the subject of much debate as to the nature of scaling laws. \citet{hoffmann-optimal} concluded data size and model size should be scaled equally to achieve compute-optimal training for LLM’s. Further credence to reducing LLM compute is lent by \citet{sorscher2022beyond}, who found input data pruning can improve models to scale exponentially, and \citet{chan2022data}, who showed LLM in-context learning derives from the Zipfian distribution of pre-training data.

Most releavant to this work is \citet{huebner2021babyberta} who found a small language model trained on child-directed speech can achieve comparable performance to larger LMs on a set of probing tasks. In contrast to them, we train multiple models, explore scaling in the low-compute region and evaluate on a filtered version of GLUE instead of a set of linguistic tests.

% Additional studies on model size have underscored the positive relationship between model size and linguistic capacity, although this relationship is also impacted by factors like model architecture \cite{van-schijndel-etal-2019-quantity}, and the domain of evaluation datasets \cite{dan-roth-2021-effects-transformer}.
% \paragraph*{Data input}
% \orange{Input data volume has also been examined for its impacts on downstream performance} Changing pre-training data size via training token volume \cite{perez-pretraining,zhang-billions}, vocabulary size \cite{gowda-may-2020-finding}, or the number of epochs \cite{Voloshina_2022} has also been explored for effects on language acquisition. These studies have generally demonstrated low-level linguistic tasks involving syntax require relatively little data volume compared to more complex tasks like WinoGrande \cite{Sakaguchi2019WINOGRANDEAA}.
% \paragraph{Scaling and simplifying language}

\section{Methodology}

This section discusses the language simplification process, pre-training data, development of the data tokenizer, language model configuration, and pre-training objective in detail.

\subsection{Simplifying language}
\label{voc}

To create a corpus of reduced English, we filter large text corpora based on a word vocabulary from AO-CHILDES \cite{huebner2021using}.
The AO-CHILDES corpus contains English transcripts of child-directed speech. With the transcripts, we generate a vocabulary by removing special characters and tokenizing words by spaces. We also remove gibberish words present in the transcripts e.g. ``bababa''. With this process, we construct a set of 21 thousand unique words.

\subsection{Pre-training data}

We filter data from five text corpora: Wikipedia, Simple English Wikipedia\footnote{\url{https://simple.wikipedia.org}}, Book Corpus \cite{Zhu_2015_ICCV}, Children's Book Test (CBT) \cite{hill2015goldilocks}, and Common Crawl (C4) \cite{raffel2020exploring}, to obtain pre-training data.
% With a length and step size of 110 and 30 whitespace-separated words, we first process the C4 corpus. We select a text span to include in the pre-training data if and only if there are no words out of the vocabulary of interest, ignoring any numeric characters. 
We filter C4 two ways: span level (110 words span size, 30 words stride) and sentence level.
We select a text span (or sentence) to include in the pre-training data if and only if there are no words out of the vocabulary of interest, ignoring any numeric characters.
% To further increase the size of the pre-training data we conduct another round of data filtration at the sentence level (instead of spans) with the same filtration rules.
For sentence-level filtration, we process text data on all five corpora. With sentence-level filtration, we collect approximately 44 million sentences which we concatenate to construct six million spans. The combination of both span- and sentence-level data filtration provided us with over nine million pre-training sequences of an average length of 127 BPE tokens. Finally, we split the filtered data into three sets: train, development, and test, of sizes nine million, 100 thousand, and 100 thousand, respectively. We provide the amount of filtered data from each text corpus in Table \ref{table:data-collection}. 
%The first row of the table for the C4 corpus refers to span-level filtering (length of 110 words and window of 30 words), and the size represents the number of spans filtered, instead of sentences. Otherwise, the numbers represent results for sentence-level filtration.

\begin{table}
\begin{small}
\centering
\begin{tabular}{l|cc}
\toprule
Corpus name & Sentences (mil.) & Tokens (mil.) \\
\midrule
C4\footnotemark{} & 3 & 427 \\
C4 & 27 & 428 \\
Book Corpus & 12 & 190 \\
Wikipedia & 4.8 & 76 \\
Simplified Wikipedia & 0.19 & 3 \\
Children's Book Test & 0.08 & 1 \\
\midrule
Total & 47.07 & 1,125 \\
\bottomrule
\end{tabular}
\caption{Size of the filtered pre-training data used in this study.% in number sentences (million) and an approximate number of BPE tokens (million) collected from various corpora.
}
\label{table:data-collection}

\end{small}
\end{table}

\footnotetext{Span-level filtering.}

For the rest of the paper, we use the word ``vocabulary'' to refer to the number of unique tokens instead of unique whitespace-separated words, unless otherwise mentioned.

\subsection{Tokenizer}

% Our collected pre-training data differs in its limited vocabulary when compared to similarly-sized text corpora. Because of the likely impact of AO-CHILDES-derived vocabulary on the vocabulary of tokens, we train the tokenizer on our data instead of using a pre-trained tokenizer.
% \todo{AR: Give one or two sentences motivating this. The motivation is, we are reducing vocabulary size, therefore we need to understand how many tokens we need.}

Since we are working with a reduced language, commonly used subword vocabulary sizes for English might be suboptimal.
% We need to determine the optimal vocabulary size.
We want to achieve a reasonable balance between over-splitting (splitting the words into units smaller than the smallest meaningful morphological units, e.g., splitting into characters) and under-splitting (e.g., retaining full words instead of splitting them into meaningful subwords). 

We conducted a series of experiments in which we tuned the vocabulary size to find the right balance.
While varying the vocabulary size, we track two metrics for the tokenized text: \textit{word-split ratio}
and another metric we define, the \textit{exact sub-token matching score} (ESMS). 
Word-split ratio is the number of tokens into which a word is split, where words are separated by whitespace. For example, if the word ``cooking'' is converted to ``cook'' and ``ing'', then the word-split ratio value is two. We measure and report the word-split ratio value for 5,000 examples sampled from the set of collected pre-training data without replacement.

%\orange{We measure and report the word-split ratio value averaged over ten samples of 5,000 sentences}\todo{Vlad: Do you mean that you trained the tokenizer on 5K sentences? Or you trained in on everything, but measured the word-split ratio on 50K? Why say 10 samples of 5K sentences instead of 50K?}. The 5,000 sentences are sampled from the set of collected nine million sentences without replacement.

To measure ESMS, we compare the tokenizer performance with morpheme-based subword tokens. For example, in case of the word ``cooking'', we check whether the tokenizer is splitting the word into two tokens, `cook' and `ing'. For this purpose, we used a manually-curated list of 127 words with their corresponding morpheme-based sub-tokens,
(see Table \ref{table:esms-word-list} in the Appendix for some examples).
ESMS is computed as an exact match to the reference tokenization. For one example, it is equal to 1 if the word is tokenized exactly as in the reference and 0 in any other case.
% We start with an ESMS value of zero and test the tokenizer on each word in the reference word list. We increase the score by one for each exact match to the reference morpheme-based token.
%e.g. if the tokenized version of 'cooking' is found to be 'cook', 'i', and 'ng' then the score will only increase by one (based on exact match of 'cook' token). 
%Moreover, we penalize the score by one if we observe a character level splitting (i.e., `c', `o', `o', `k', `i', `n', `g') or no splitting at all (i.e., `cooking'). \todo{Vlad: this needs to be improved. Very hard to read.} The maximum and minimum values of ESMS that can be achieved with our reference dictionary are 27 and -13, respectively, with a set of 13 words.
% After we process all words in the reference dictionary, we normalize the final score by the number of maximum possible exact matches for our reference word list. Hence, the ESME value ranges between 0 to 1, representing desirable tokenizers for higher values.

We experiment with three types of tokenizers, Byte-Pair Encoding (BPE) \cite{radford2019language}, WordPiece \cite{devlin2018bert}, and SentencePiece \cite{raffel2020exploring}. Similar to the study conducted by \citet{fitzgerald2022alexa}, we select vocabulary size for each type of tokenizer by minimizing the absolute difference of word-split ratio compared to the reference tokenizer. We consider separate reference tokenizers for each tokenizer type. For BPE, WordPiece, and SentencePiece, we select pre-trained tokenizers published by \cite{liu2019roberta}, \cite{devlin2018bert}, and \cite{raffel2020exploring}, respectively, as our reference tokenizers. After selecting the vocabulary size for each tokenizer type, we select the tokenizer with the highest value of ESMS as our final choice. 

%In other words, we sequentially applied two rules for tokenizer selection: first, minimization of the difference between word-split ratio, and second, maximization of ESMS value. \orange{The first rule selects a size of the vocabulary that can minimize over- and under-splitting for the entire pre-training data. While the second rule further penalizes over- and under-splitting by considering specific cases present in our pre-training data, hence, finding the best-suited tokenizer for our application case.}\todo{Vlad: I do not understand what you mean by first rule maximizes this, then second rule minimizes this.}

With the above-mentioned selection process, we find that the BPE tokenizer with a vocabulary size of 19,000 and ESMS of 0.2604, is the best-suited tokenizer for our study. We provide the results of our tokenizer selection experiments in Appendix~\ref{appendix:tokenizer}.

\subsection{Model architecture and configuration}
\label{subsection:model-config}

The models we pre-train in our experiments closely follow the configuration setting of RoBERTa \cite{liu2019roberta}. We scale down and vary the model's hidden size, intermediate size (FFN hidden dimension size), number of hidden layers, and number of attention heads such that the total number of trainable parameters does not exceed 20 million. To separately control model hidden size and embedding size, we also add a linear layer, followed by a normalization layer \cite{ba2016layer} between the embedding block and the first Transformer layer.
% Moreover, we add one linear layer followed by a normalization layer \cite{ba2016layer} between the embedding block and the first hidden layer of our language model. The primary motivation behind adding a linear layer is to facilitate changes in the embedding and the hidden vector dimension, independent of each other.

\subsection{Pre-training objective}

% In our study, we pre-train models on a Masked Language Modeling (MLM) task. We employ a \red{simplified version of the MLM objective}\todo{Vlad: what do you mean by this? Ok, I read the rest and understood it. This section needs to be rephrased.} used for pre-training BERT \cite{devlin2018bert}. For pre-training BERT, the authors of the study used a 15\% masking strategy i.e., 15\% tokens in the input sequence will be selected for masking. Out of all tokens selected for masking, 80\% of the tokens were replaced with the mask token $'<mask>'$, 10\% tokens were replaced by a random token from the vocabulary, and 10\% were kept the same. 

% We conducted an exploratory set of experiments to observe the effect of various MLM objective settings on evaluation loss and perplexity. We found, especially for smaller models ($\leq 10\ million$), that \red{evaluation metrics}\todo{Vlad: upstream or downstream metrics?} reached a plateau of a significantly higher value when using the 10\% random and 10\% same word replacement, compared to 0\% random and 0\% same word replacement strategy. Hence, to enable considerable learning in the limited parameter setting we implement 0\% random replacement and 0\% same-word replacement of the token selected for masking. In other words, we always replace the token selected for masking with the mask token $<mask>$, before inputting it into the model. We otherwise adopt the same strategy as BERT pre-training by masking 15\% of tokens.

% new version of the section
In our study, we pre-train models on a Masked Language Modeling (MLM) task \cite{devlin2018bert}. We chose MLM instead of regular (causal) language modeling, because of its effectiveness for natural language understanding tasks at a smaller scale as demonstrated by BERT.

We conducted an exploratory set of experiments to observe the effect of various MLM objective settings on validation perplexity. We found that using a random word replacement strategy and same-word replacement strategy doesn't improve the model at a small scale. Hence, to enable considerable learning in the limited parameter setting, we do not use random replacement and same-word replacement of the token selected for masking. In other words, we always replace the token selected for masking with the mask token \texttt{<mask>} before inputting it into the model. Otherwise, we adopt the same strategy as BERT pre-training by masking 15\% of tokens.

\section{Experimental Setup}

In our experiments, we explore the relationship between training data size, model size (number of parameters), model shape, cost of training (FLOPs), and performance during pre-training and downstream. 
%Such exploration allowed us to study not only the effect of model size but also the effect of model shape on various cost and performance measurements. 
In the following subsections, we will discuss our strategy for exploring various model shapes followed by a discussion on hyperparameter settings in detail.

\subsection{Exploration of model configuration}

To investigate the impact of reduced model size, we start by scaling down the base configuration of RoBERTa \cite{liu2019roberta} from its initial hidden size of 768 to 256, and the number of hidden layers and attention heads from 12 to 8. For intermediate layer size, we follow the common practice of setting it to a value four times that of the hidden size. We refer to this configuration as the anchor configuration. We pre-train a model with the anchor configuration and explore three values for embedding size, hidden size, intermediate size, number of hidden layers, and number of attention heads, varying one hyperparameter at a time. With such unidirectional exploration, we pre-train 16 models. We refer to this set of 16 models as \textbf{set-1}. To explore more model configurations, we randomly sample 16 configurations that are not included in \textbf{set-1}. For random sampling, we only explore values that are powers of two and are upper-bounded by 256, 256, 1024, 8, and 8, for the embedding size, hidden size, intermediate size, number of attention heads, and number of hidden layers, respectively. We refer to this set of 16 models as \textbf{set-2}. Furthermore, we pre-train 30 more models by performing unidirectional explorations of hidden size and the number of hidden layer values by anchoring other hyperparameter values. We refer to this set of 30 models as \textbf{set-3}.
%The full list of model configurations is available in Appendix \ref{appendix:model-config-list}.

% For the cost-effectiveness analysis, we utilize models in \textbf{set-1}. Although, for the curve-fitting analysis we utilize all pre-trained models. For downstream evaluations, we primarily fine-tune the \textbf{set-1} models but we also sample more models from \textbf{set-2} and \textbf{set-3} for fine-tuning.

\subsection{Pre-training} 
\label{pt-ex}
% \todo{Vlad: reduce this section and move params to appendix when we hit the page limit}

For every model configuration, we keep the input sequence length fixed at 128 tokens.
% According to the availability of computing resources, we adjust the gradient accumulation steps and fix the effective batch size value at 256 input sequences, for all models.
% Prior to training, we initialize the weight parameters of the model with Xavier initialization \cite{glorot2010understanding} and bias parameters to a value of $0.01$.
% We use the values 1 and 0 for initialization for the layer normalization weight and bias parameters.
All models are initialized with a fixed random seed value of zero. 

Once initialized, we train the model for one epoch with a batch size of 256 for 35,000 weight updates.
% , without data duplication. Over the training horizon of approximately 35k updates, we schedule the learning rate similar to NOAM \cite{vaswani2017attention}, implementing a version that is independent of model size. Precisely, our implementation is as follows,
We use an inverse square root learning rate scheduler with 5\% warmup.
% \begin{equation}
%     lrate = min(step\_num^{-0.5}, step\_num \cdot warmup\_steps^{-0.5}) 
% \end{equation}
% We fix the warmup period of the scheduler to 5\% of the total number of parameter updates, yielding approximately 1,750 updates.
% To choose a peak learning rate value, we first followed the guide provided by \cite{kaplan2020scaling} i.e.,
% \begin{equation}
%     LR(N) = 0.003239 + (-0.0001395 \cdot log(N))
% \end{equation}
% where, $N$ is the number of non-embedding parameters in the model. Although, we observed instability in the training with the values suggested by the above formula. Hence, we reduced the formula-based learning rate values until we observed a stable loss curve. By this procedure, we set the peak learning rate of $1 \times 10^{-4}$ for all models trained under the anchor configuration having the highest hidden size of 512. Otherwise, i.e., for smaller models, we use a larger value of peak learning rate, $5 \times 10^{-4}$.

%We chose the peak learning rate of $5 \times 10^{-4}$ for most of our pre-trained models. Although, for a few model configurations explored in \textbf{set-3}, we set the peak learning rate to $1 \times 10^{-4}$ due to the considerably larger size.

We conducted a few trials guided to decide the peak learning rate value for our experiments. We started with values higher than 6e-4, based on findings published by \citet{liu2019roberta} and \citet{kaplan2020scaling}, and kept on reducing the learning rate until we observed a stable training loss curve. We observed $1 \times 10^{-1}$ to be suitable for models with more than 18 million parameters and $5 \times 10^{-1}$ otherwise.

For optimization, we use the AdamW optimizer \cite{loshchilov2017decoupled} with $\beta_1$,  $\beta_2$, and $\epsilon$ values set to 0.9, 0.95, and $10^{-8}$, respectively. After preliminary experiments we set the weight decay parameter to 0.01. Besides the learning rate, we keep all optimizer-related hyperparameters constant across all model configurations. For all dropout layers in the model, we adopt the same value of 10\% as that of the RoBERTa model \cite{liu2019roberta}.

% Throughout pre-training, we track cross-entropy loss and perplexity. At every $1000^{th}$ parameter update, we evaluate the model on the development set of 100,000 sequences with the MLM task. For every improvement in development set perplexity, we save (or update) the checkpoint corresponding to the best model. At the end of pre-training, we load the checkpoint corresponding to the best model and measure the performance of the best model on the test set (100,000 unseen examples) for the MLM task.

%We report the cross-entropy loss and perplexity on the test split of the data. Nevertheless, we leverage the cross-entropy loss and perplexity calculated on the development set for part of our analysis.

\subsection{Fine-tuning} \label{ft-ex}
We evaluate pre-trained models on GLUE \cite{wang2018glue}.
% various natural language understanding tasks included in The General Language Understanding Evaluation (GLUE) benchmark dataset \citet{wang2018glue}.
Because our pre-trained data consists of a limited vocabulary, we fine-tune and test on GLUE task datasets with the same vocabulary filtering, in addition to unfiltered variants.

For all tasks, we fine-tune our pre-trained models for 5 epochs and report the performance of the best performance value on the validation set averaged over three seed values. For all fine-tuning experiments, we keep the batch size fixed at 32. Over the five training epochs, we vary the learning rate value with a linear scheduler with a warmup of 5\%. We set the peak-learning rate within the range from 2e-5 to 2e-4 value, according to the task. In addition to these pre-trained models, we fine-tune and evaluate GLUE for randomly-initialized versions of pre-trained models as well.

\subsection{Evaluation metrics}

\paragraph{Pre-training}
For pre-training results, we measure and report the cross-entropy loss and perplexity on the test split of the data. We use the cross-entropy loss and perplexity calculated on the development set for curve fitting. 
In both cases, we calculate the cross-entropy loss only for the masked tokens, and the perplexity value is calculated by exponentiating the cross-entropy loss value.
%part of our analysis (curve-fitting). 
We also calculate the FLOPs (compute cost) as defined by \citet{hoffmann-optimal}. We first calculate FLOPs per training sequence based on the model parameters (including the embedding parameters) and multiply it by the amount of training data seen by the model to get a total number of FLOPs. We provide a detailed formula of FLOPs calculation in Appendix \ref{appendix:FLOPs-calculation}.

\paragraph{Cost-effectiveness analysis}
We use the Incremental Cost-Effectiveness Ratio (ICER) \cite{bambha2004cost} to conduct cost-effectiveness analysis of different model configuration hyperparameters. We treat the FLOPs and model perplexity values as proxies for the expenditure and the outcome of expenditure, respectively. Therefore,

\begin{equation}
    ICER = \frac{\Delta Outcomes}{\Delta Cost} = \frac{\Delta Perplexity}{\Delta FLOPs}
\end{equation}

We calculate the difference (the $\Delta$ values) by comparing a model configuration with the next cheaper option (e.g., we compare the model with a hidden size of $2^{8}$ to the model with a hidden size of $2^{7}$). For the specific case of increasing hidden size from $2^{7}$ to $2^{8}$, ICER represents performance gain (reduction in perplexity) per additional FLOPs spent on increasing the hidden size value from $2^{7}$ to $2^{8}$.
We calculate ICER values for four hyperparameters namely, embedding size, hidden size, intermediate size, and the number of hidden layers.

%While exploring the different model configurations for pre-training we varied various configuration hyperparameters one at a time e.g., we increase the hidden size from $2^{5}$ to $2^{8}$ by keeping all other hyperparameters constant. Each unidirectional increment increases the number of parameters and hence, FLOPs, that directly correspond to the varied hyperparameter. Likewise, every unidirectional increment also holds an effect on the loss and the perplexity values that are directly associated with the varied hyperparameter. Hence, by arranging the hyperparameter values from lowest to highest we calculate the addition in the FLOPs and the change in perplexity for each increment. By treating the FLOPs values as a proxy for cost and perplexity values as the outcome of the cost, we calculate Incremental Cost-Effectiveness Ratio (ICER) \cite{bambha2004cost}. 

\paragraph{Fine-tuning}
We use standard metrics for the GLUE benchmark: accuracy, Matthew's correlation score, and combined correlations score depending on the task.
% We focus our fine-tuning experiments on the GLUE benchmark dataset that includes nine natural language understanding tasks. For the linguistic acceptability task, we measure Matthew's correlation score. For the textual similarity task, we calculate the combined score i.e., the average value of Pearson and Spearman correlation coefficient. For all other tasks, we resort to the accuracy values.
For our conducted experiments, we report the average value of all performance metrics across tasks.

\section{Results and Discussion}

% \begin{figure}[ht!]
% \centering
% \includegraphics[width=75mm]{final_tiny_glue_hp_flops.png}
% \caption{Pre-trained and fine-tuned model GLUE performance relative to randomly initialized fine-tuned baselines. L represents number of hidden layers.}
% \label{figure:FT-GLUE-hidden-layers}
% \end{figure}

\subsection{Curve fitting}

To assess the empirical relationship between model performance and data size, model size, and FLOP values,
% we conduct a curve-fitting analysis. We fix our response variable ($y$) to the MLM loss value and
we fit a power curve of the form $y = C \cdot x^{e}$, separately, to model size, data size, and FLOP values. We only consider the compute-optimal instances for curve fitting. To find compute optimal instances, we first divide the FLOPs values into over 30 bins and fetch the checkpoint corresponding to the minimum value MLM loss for each FLOPs bin.
We use an implementation of the Levenberg-Marquardt \cite{more1978levenberg} algorithm provided under the $SciPy$ library for curve fitting.

\begin{figure}[t]
\centering
\includegraphics[width=75mm]{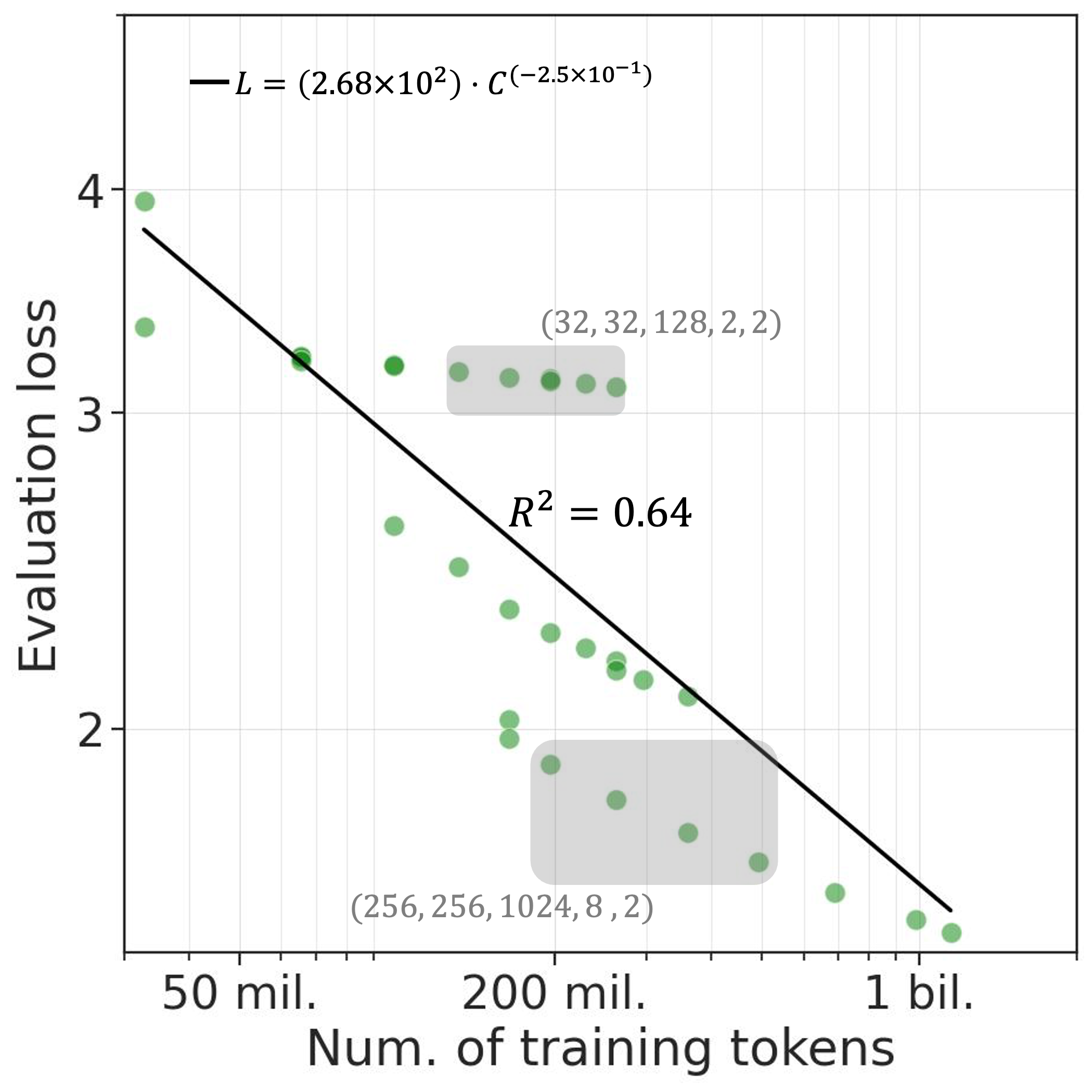}
\caption{\textbf{Reduction of MLM loss for increasing size of pre-training tokens.} The green dots represent the compute-optimal instances found in our experiments. The solid black line represents the fitted power curve with the exact equation in the top left corner. The gray highlighted regions are examples of regions where we find a model configuration (of the format (embedding size, hidden size, intermediate size, number of layers, number of attention heads)) consistently being compute-optimal. }
\label{figure:PT-data-vs-loss}
\end{figure}

\begin{figure}[t]
\centering
\includegraphics[width=75mm]{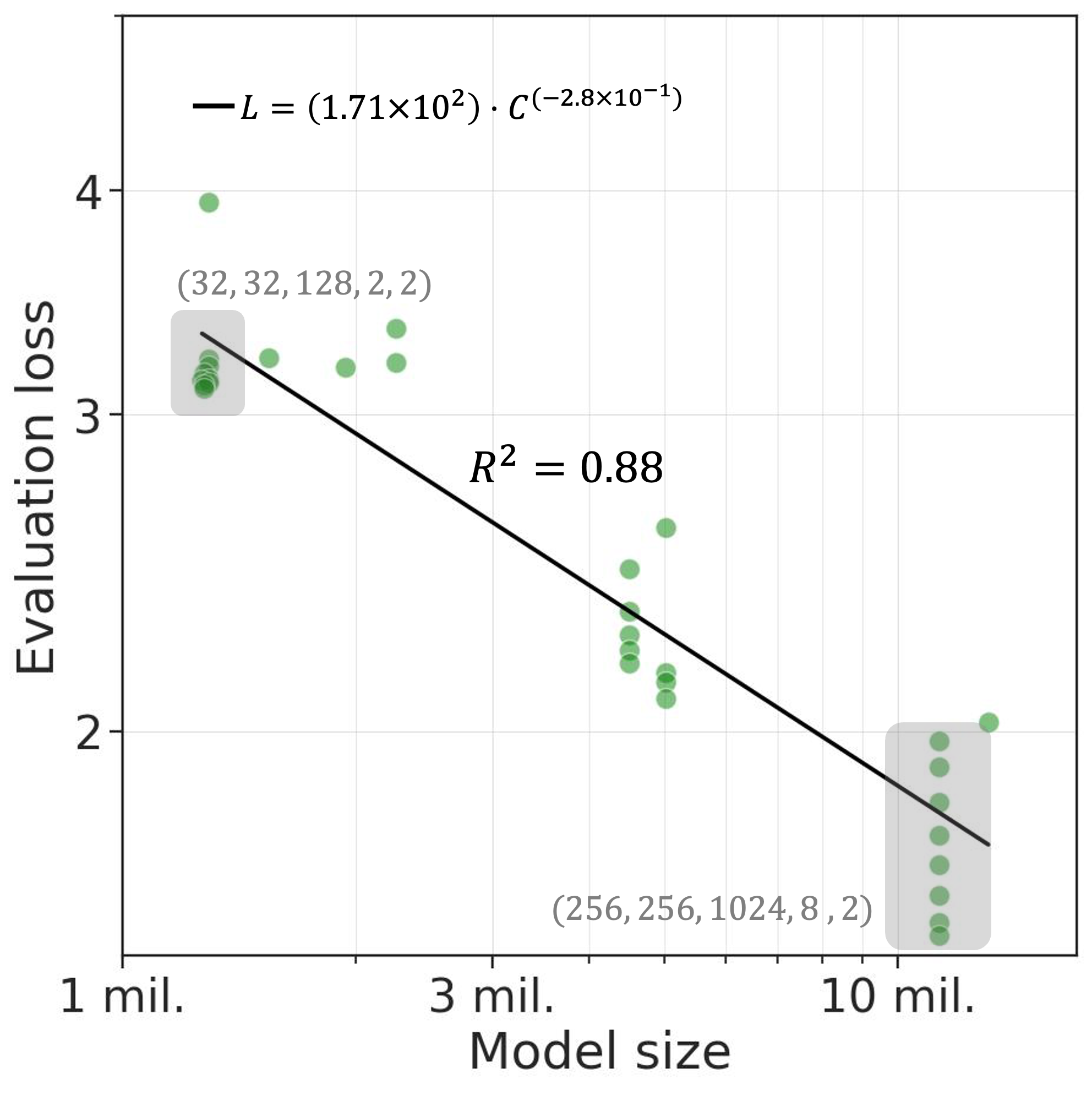}
\caption{\textbf{Reduction in the MLM loss for increasing model size.} The green dots represent the compute-optimal instances found in our experiments. The solid black line represents the fitted power curve with the exact equation in the top left corner. The gray highlighted regions are examples of regions where we find a model configuration (of the format (embedding size, hidden size, intermediate size, number of layers, number of attention heads)) consistently being compute-optimal.}
\label{figure:PT-model-size-vs-loss}
\end{figure}

We observe that the optimal values of the exponents for data size and model size are, $-0.2459$ and $-0.2805$. Note that these values are expected to be different from those in \citet{kaplan2020scaling}, since we work with a different loss and a reduced-vocabulary language. 
%(cf. Table \ref{table:result-exponents}). 
The small difference between both exponent values suggests that the MLM loss reduces with a similar pace for data and model scaling. Hence, in our downscaled problem, we find data and model scaling equally important for compute-optimality. Although, we find $R^2$\footnote{$R^2$ is the coefficient of determination and we adopt the default definition of $R^2$ in the Scipy Python library. Please refer to \hyperlink{https://scikit-learn.org/stable/modules/generated/sklearn.metrics.r2_score.html}{Scipy} for more details.} values for both curves i.e., loss vs. data size and loss vs. model size, are low (Figures \ref{figure:PT-data-vs-loss} and \ref{figure:PT-model-size-vs-loss}). Moreover, we observe that for FLOPs values greater than $2\time10^{15}$, a power curve nearly perfectly predicts the compute optimal MLM loss value for a given compute budget. In this region, we find the exponent for the FLOPs values to be $-0.1412$ with an $R^2$ value of $0.9888$. 

In our experiments, we find few model configurations to be effective for a long range of FLOP values. We highlight a couple of examples of such configurations in Figures \ref{figure:PT-data-vs-loss} and \ref{figure:PT-model-size-vs-loss}. The occurrence of such a configuration causes a discontinuous transition of MLM loss with respect to data size, model size, and FLOP values. We observe this effect to be more pronounced in the lower FLOPs region. Consequentially, a power curve does not fit in the region of lower FLOP values ($\leq 2 \times 10^{15}$). With the exponent value of $-0.0929$, we observe the best value of $R^2$ to be $0.68$.

In order to illustrate the effects of parameter count increase in the downscaled setting, we extended our pre-training experiments to include models of larger size in Figures \ref{figure:PT-data-vs-loss} and \ref{figure:PT-model-size-vs-loss}. We observed that increasing the parameter count up to ~100 million does not allow the model to beat the perplexity achieved by a smaller 16 million parameter model (See Appendix \ref{appendix:larger-models}).

%We extended our pre-training experiments to include models of larger size in Figures \ref{figure:PT-data-vs-loss} and \ref{figure:PT-model-size-vs-loss}, however we find that we lack sufficient training data to achieve model perplexity that corresponds with compute optimal performance \ref{table:appendix-larger-models}.

% \begin{table}[h!]
% \centering
% \begin{tabular}{lcc}

% \toprule

% \textbf{$L_{opt} \propto$} & \textbf{$D^{exponent}$} & \textbf{$N^{exponent}$} \\

% \midrule

% \citet{kaplan2020scaling} & -0.0950 & -0.0760 \\
% Ours & -0.2459 & -0.2805 \\
% \bottomrule

% \end{tabular}

% \caption{Power curve exponent values calculated from the compute-optimal runs in our experiments. $L_{opt}$ is the MLM loss value corresponding to a compute-optimal instance, $D$ is the number of training tokens and $N$ is the model size (inclusive of the embedding parameters). }
% \label{table:result-exponents}
% \end{table}

% \begin{table}
% \centering
% \begin{tabular}{lll}

% \toprule

% \multicolumn{1}{p{2cm}}{\centering \textbf{}}
% & \multicolumn{1}{p{2cm}}{\centering \textbf{Data size}}
% & \multicolumn{1}{p{2cm}}{\centering \textbf{Model size}} \\

% \midrule

% \multicolumn{1}{p{2cm}}{\centering \citet{kaplan2020scaling}}
% & \multicolumn{1}{p{2cm}}{\centering -0.0950}
% & \multicolumn{1}{p{2cm}}{\centering -0.0760} \\

% \multicolumn{1}{p{2cm}}{\centering Ours}
% & \multicolumn{1}{p{2cm}}{\centering -0.2459}
% & \multicolumn{1}{p{2cm}}{\centering -0.2805} \\

% \bottomrule

% \end{tabular}

% \caption{Power curve exponent values computed from the compute-optimal runs in our experiments}
% \label{table:result-exponents}
% \end{table}

\subsection{Incremental cost-effectiveness analysis}

For the cost-effectiveness analysis, we focus on the pre-trained model configurations in \textbf{set-1}. We calculate the ICER values separately for four hyperparameters: embedding size, hidden size, intermediate size, and the number of hidden layers. We arrange the models in increasing order of FLOPs and calculate ICERs by comparing each model to the next cheapest option. The ICER values presented in Table \ref{table:results-ICER} represent performance gain (reduction in perplexity) per an additional expenditure of a billion FLOPs, scaling only one hyperparameter at a time.

We observe the highest ICER value of $3.0075$ for scaling the hidden size from 32 to 64, refer to Table \ref{table:results-ICER}. For further scaling of hidden size, from 64 to 128, and from 128 to 256, ICER values drop at least by 3x for each increment. Besides rapidly reducing values, ICERs for hidden size were always the highest, making it the most cost-effective choice. Comparably high ICERs were observed for scaling the model by increasing the number of hidden layers. For increasing the hidden layers from one to two, we record an ICER of $2.6271$. This value reduces to $0.6810$ and $0.2089$, when scaling the model with two layers to have four layers, and a model with four layers to have eight layers, respectively. 

We find the ICER values for embedding size and intermediate size significantly lower than the values for hidden size and the number of hidden layers. The differences between ICERs were higher for the lower values of each hyperparameter. Although, when all hyperparameter values reached the corresponding highest values, the difference in ICERs diminished. A comparison between ICER values for embedding size and intermediate size shows that increasing embedding size from 32 to 64 brings $0.2275$ more improvement in the perplexity per million FLOPs, compared to increasing intermediate size from 128 to 256. However, for all further increments in the hyperparameter values, increasing intermediate size results in at least $0.03$ more ICER value than for embedding size.

\begin{table}
\centering
\begin{small}
\begin{tabular}{lll}

\toprule

\textbf{Model config. (E, H, I, L, A)}
& \textbf{ICER} \\

\midrule
(256, \textbf{32}, 1024, 8, 8) & -- \\
(256, \textbf{64}, 1024, 8, 8) & 3.0075 \\
(256, \textbf{128}, 1024, 8, 8) & 0.8316 \\
(256, \textbf{256}, 1024, 8, 8) & 0.2411 \\

\midrule
(256, 256, 1024, \textbf{1}, 8) & -- \\
(256, 256, 1024, \textbf{2}, 8) & 2.6271 \\
(256, 256, 1024, \textbf{4}, 8) & 0.6810 \\
(256, 256, 1024, \textbf{8}, 8) & 0.2089 \\

\midrule
(\textbf{32}, 256, 1024, 8, 8) & -- \\
(\textbf{64}, 256, 1024, 8, 8) & 0.6277 \\
(\textbf{128}, 256, 1024, 8, 8) & 0.2105 \\
(\textbf{256}, 256, 1024, 8, 8) & 0.1669 \\

\midrule
(256, 256, \textbf{128}, 8, 8) & -- \\
(256, 256, \textbf{256}, 8, 8) & 0.4002 \\
(256, 256, \textbf{512}, 8, 8) & 0.4127 \\
(256, 256, \textbf{1024}, 8, 8) & 0.1970 \\

\bottomrule

\end{tabular}

\caption{Incremental cost-effectiveness ratio values for increasing values of various hyperparameters. Model hyperparameters map accordingly,  E: embedding size, H: hidden size, I: intermediate size, H: number of hidden layers, A: number of attention heads. Because we calculate ICER values relative to the model with the next lowest value of hyperparameter, ICERs are not calculated for the lowest values of hyperparameters.}
\label{table:results-ICER}
\end{small}
\end{table}

\begin{table*}[th!]
\begin{small}
\centering
\begin{tabular}{ccccccc}

\toprule

  \textbf{\makecell{Model config. \\ (E, H, I, L, A)}}
& \textbf{\makecell{Model size \\ (mil. parameters)}}
& \textbf{\makecell{FLOPs \\ ($\times 10^{15}$)}}
& \textbf{Perplexity}
& \textbf{\makecell{GLUE score \\ (unfiltered)}}
& \textbf{\makecell{GLUE score \\ (filtered)}}
& \textbf{\makecell{GLUE score \\ (filtered, w/o PT)}} \\

\midrule

(256, 256, 1024, 8, 8) & 16.24 & 110 & 4.80 & 51.73 & 56.29 & 40.24 \\

\midrule

(\textbf{32}, 256, 1024, 8, 8) & 11.89 & 80 & 5.60 & 46.99 & 48.39 & 49.80 \\
(\textbf{64}, 256, 1024, 8, 8) & 12.51 & 84 & 5.31 & 45.12 & 51.09 & 51.99 \\
(\textbf{128}, 256, 1024, 8, 8) &  13.75, & 92 & 5.11 & 48.07 & 52.73 & \textbf{52.09} \\

\midrule

(256, \textbf{32}, 1024, 8, 8) & 6.10 & 42 & 10.42 & 47.98 & 50.23 & 45.93 \\
(256, \textbf{64}, 1024, 8, 8) & 7.34 & 50 & 7.56 & 49.34 & 53.78 & 50.67 \\
(256, \textbf{128}, 1024, 8, 8) & 10.04 & 69 & 5.88 & 51.16 & 55.63 & 50.60 \\
% (256, \textbf{256}, 1024, 8, 8) &16.24 &110 &4.80 &51.73 & 56.29 & 40.24 \\

\midrule

(256, 256, \textbf{128}, 8, 8) &7.63 & 85 & 5.61 &57.65 & 57.18 & 41.62 \\
(256, 256, \textbf{256}, 8, 8) & 8.15 & 88 & 5.45 & 54.91 & 56.48 & 41.28 \\
(256, 256, \textbf{512}, 8, 8) & 9.20 & 96 & 5.12 & 51.60 & 57.36 & 40.55 \\
% (256, 256, \textbf{1024}, 8, 8) & 16.24 & 110 & 4.80 & 51.73 & 56.29 & 40.24 \\

\midrule

(256, 256, 1024, \textbf{1}, 8) & 10.71 & 73 & 7.60 & 50.87 & 55.99 & 50.53 \\
(256, 256, 1024, \textbf{2}, 8) & 11.50 & 79 & 6.07 & 54.31 & 59.39 & 51.17 \\
(256, 256, 1024, \textbf{4}, 8) & 13.08 & 89 & 5.28 & 53.85 & \bf 60.67  & 47.15 \\
% (256, 256, 1024, \textbf{8}, 8) & 16.24 & 110 & 4.80 & 51.73 & 56.29 & 40.24 \\

\midrule

(256, 256, 1024, 8, \textbf{1}) & 16.24 & 110 & 4.74 & 50.47 & 53.68 & 40.30 \\
(256, 256, 1024, 8, \textbf{2}) & 16.24 & 110 & \textbf{4.67} & 50.39 & 54.98 & 40.10 \\
(256, 256, 1024, 8, \textbf{4}) & 16.24 & 110 & 4.75 & 49.55 & 54.21 & 39.66 \\
% (256, 256, 1024, 8, \textbf{8}) & 16.24 & 110 & 4.80 & 51.73 & 56.29 & 40.24 \\

\midrule

(32, 32,128, 2, 2) & 1.27 & 8.57 & 20.07 & 46.25 & 49.03 & 48.68 \\
(32, 32, 128, 1, 1) & 1.25 & 8.60 & 23.40 & 44.97 & 48.22 & 47.98\\
(32, 32, 64, 1, 1) & 1.25 & 8.71 & 23.42 & 44.91 & 47.20 & 48.69 \\

\bottomrule

\end{tabular}

\caption{Performance on vocabulary-filtered and unfiltered GLUE benchmarks. These models have access to the same number of tokens during pre-training. Model hyperparameters map accordingly: Embedding Size: E, Hidden Size: H, Intermediate Size: I, Hidden Layers: L, Attention Heads: A. In the last column, we report the fine-tuning results for the randomly initialized (without pre-training) counterpart of the model.}
\label{table:result-GLUE}
\end{small}
\end{table*}

\subsection{Downstream evaluation}

\begin{figure}[ht!]
\centering
\includegraphics[width=75mm]{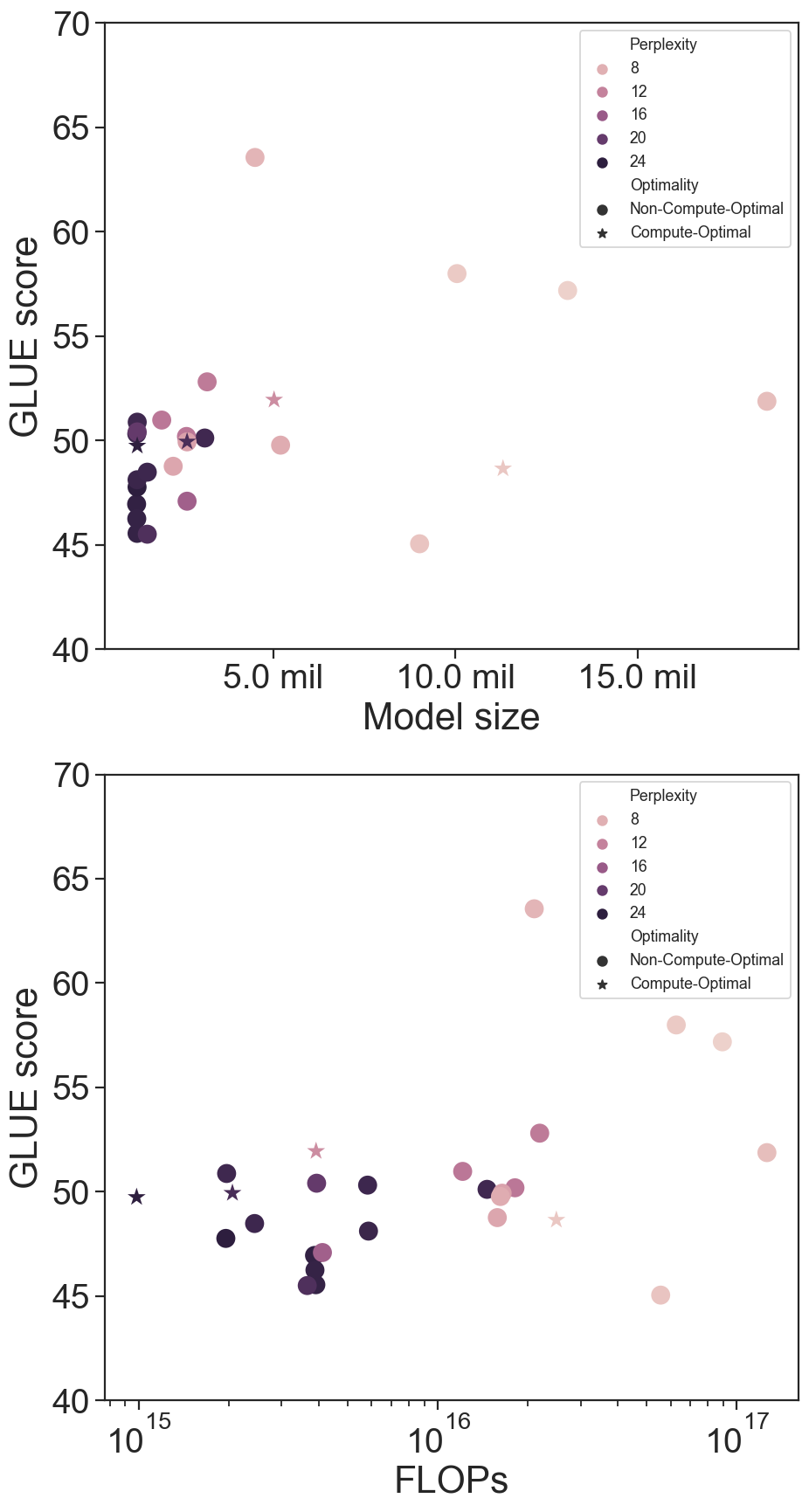}
\caption{GLUE scores by model size and FLOPs for compute-optimal and non-compute-optimal models}
\label{figure:FT-GLUE-compute-optimal}
\end{figure}

We report fine-tuning performance on vocabulary-filtered GLUE benchmark in Table \ref{table:result-GLUE} (cf. Section \ref{voc}). For reference, we also report performance on unfiltered GLUE. We find that GLUE performance peaks for models with 2 and 4 hidden layers with average scores of 59.39 and 60.67, respectively. Interestingly, we find the average GLUE score decreases for the model with 8 hidden layers to 56.29. Such reduction is not observed when increasing the hidden size or the embedding size. As expected, models consistently demonstrate better performance on vocabulary-filtered GLUE. We also see that model performance is strongest for models with 2 and 4 hidden layers assessed on vocabulary-filtered GLUE.

To assess whether pre-training effects are beneficial in the downscaled setting,  we compare the average GLUE score of each pre-trained model with the score of the same model fine-tuned without pre-training. Table \ref{table:result-GLUE} shows that for each model shape, the fine-tuned model outperforms its respective randomly initialized counterpart. 
%This difference in performance indicates the capacity of downscaled models to encode general language capacity during pre-training, improving downstream GLUE performance by as much as 16 points.
Our results show that in a reduced-vocabulary setting, the advantages of pre-training are observable even for smaller models, starting with a 1.25M parameter count.

We further fine-tune a set of 27 models, comprising a mix of compute-optimal and non-compute-optimal checkpoints, to better understand the relation between upstream and downstream performance.
%To further understand this relationship between model scaling, and GLUE performance, we fine-tune a group of 27 compute-optimal and non-compute-optimal models. 
In Figure \ref{figure:FT-GLUE-compute-optimal}, we plot each model's GLUE score against its size and number of FLOPs, with color indicating the test perplexity of each model. We observe that for a given parameter count, compute-optimal models do not necessarily outperform the undertrained models on the GLUE benchmark.

Lastly, considering all fine-tuning results together, we conduct a test to measure the correlation between perplexity (upstream performance) and average GLUE score (downstream performance). We find that the correlation between the average GLUE score for unfiltered GLUE datasets and model perplexity is inconclusive with the Spearman coefficient value of -0.17 and a p-value of 0.28. On the other hand, we find average GLUE score calculated for filtered GLUE datasets highly correlates with model perplexity with the Spearman coefficient value of -0.67 and a p-value $\leq 0.01$.\footnote{Our reported values of the test metric are calculated over a sample size of 32. The exact p-value is $6.25\times10^{-7}$.}

\begin{table*}[th!]
\begin{small}
\centering
\begin{tabular}{cccccc}

\toprule

  \textbf{\makecell{Model config. \\ (E, H, I, L, A)}}
& \textbf{\makecell{$\Delta$ Model size \\ (mil. parameters)}}
& \textbf{\makecell{\% $\Delta$ \\ FLOPs}}
& \textbf{\makecell{\% $\Delta$ \\ Perplexity}}
& \textbf{\makecell{\% $\Delta$ GLUE score \\ (filtered)}}
& \textbf{\makecell{\% $\Delta$ GLUE score \\ (unfiltered)}} \\

\midrule

(256, 256, 1024, 8, 8) & 5.13 & 32.87 & 23.96 & -18.28 & -11.79 \\

\midrule

(256, \textbf{32}, 1024, 8, 8) & 2.89 & 47.59 & 17.86 & -3.58 & 2.15 \\
(256, \textbf{64}, 1024, 8, 8) & 3.21 & 44.01 & 16.16 & -3.32 & 1.60 \\
(256, \textbf{128}, 1024, 8, 8) & 3.85 & 38.99 & 20.85 & -4.97 & -0.08 \\
% (256, \textbf{256}, 1024, 8, 8) &16.24 &110 &4.80 &51.73 & 56.29 & 40.24 \\

\midrule

(256, 256, 1024, \textbf{1}, 8) & 5.13 & 49.44 & 18.90 & -15.39 & -1.90 \\
(256, 256, 1024, \textbf{2}, 8) & 5.13 & 46.12 & 17.24 & -18.99 & -7.80 \\
(256, 256, 1024, \textbf{4}, 8) & 5.13 & 40.66 & 16.93 & -19.93 & -7.61 \\
% (256, 256, 1024, \textbf{8}, 8) & 16.24 & 110 & 4.80 & 51.73 & 56.29 & 40.24 \\

\midrule

(32, 32, 128, 2, 2) & 0.65 & 51.23 & 18.90 & -8.69 & -1.00 \\
(32, 32, 128, 1, 1) & 0.65 & 51.89 & 25.79 & -9.06 & 1.19 \\
(32, 32, 64, 1, 1) & 0.65 & 52.07 & 14.78 & -8.94 & -2.05 \\

\bottomrule

\end{tabular}

\caption{Relative performance of the models trained on text data without any vocabulary filtration (unconstrained case). In this table, we present the results relative to the same configuration model trained on limited-vocabulary language data (constrained case). The $\Delta$ symbol represents the difference {$\Delta = value_{unconstrained} - value_{constrained}$} in the results and $\% \Delta$ represents the percentage difference {$\Delta = 100 \times \Delta / value_{constrained}$}. 
Perplexity values are calculated on the test split of constrained data.
Positive values for $\%\Delta$ perplexity mean a degradation of performance for the unconstrained case. Note that for the unfiltered GLUE, we do not expect limited-vocabulary models to do well.
However, we observe that besides adding a compute overhead, pre-training language models on unconstrained data degrade the pre-training and downstream performance in the models with $\leq 22$ million parameters.}
\label{table:appendix-constrained-vs-unconstrained}
\end{small}
\end{table*}

\subsection{Comparison with unconstrained text}

% AR: smaller models are needed for downscaled language

To highlight the ability of smaller models to benefit from pre-training when the language size is reduced (rather than unconstrained), we also pre-trained 10 models on unconstrained language (i.e., without any vocabulary reduction). We provide the details of the data collection, tokenizer training, and the experimental setup in Appendix \ref{appendix:comparison-with-unconstrained}. 
%We provide results from this set of experiments in Table \ref{table:appendix-constrained-vs-unconstrained}.

Table \ref{table:appendix-constrained-vs-unconstrained} shows the relative performance figures for the models trained on unconstrained language.  We fix the model configuration and report the change in performance relative to the constrained (limited-vocabulary) case. The training data size and hyperparameter setting for pre-training and fine-tuning are kept the same.

Note that for the unconstrained case, there is a considerable increase in the model size due to an increase in the Byte-BPE vocabulary (from 19,000 to 29,000). The increased model size also increases the compute cost by at least 32.87\%. Despite the increased model size and compute cost, no corresponding improvement in pre-training performance is observed, as shown in Table \ref{table:appendix-constrained-vs-unconstrained}. In fact, although the perplexity on reduced-vocabulary data decreases with model size, none of the model configurations studied reach the MLM perplexity of the reduced-scale model, when evaluated on the test split of the data. 

Since perplexity values are directly impacted by the increase in the Byte-BPE vocabulary size, we also evaluate the unconstrained data models on GLUE benchmarks for fairer evaluation.
Similar to pre-training results, we observe a consistent degradation of the performance for the filtered versions of the GLUE benchmark. For all model configurations considered, the average GLUE score for filtered datasets reduces by up to $\approx 20\%$ due to pre-training on unconstrained data. On the other hand, for the unfiltered version of the GLUE datasets, we do not expect models trained in limited-vocabulary data to do well. However, we find that the average GLUE score on unfiltered datasets improves only in three of the 10 model configurations we considered.
These results further confirm that limiting the vocabulary benefits the models with $\leq 22$ million parameters.

%Therefore, from the comparative analysis results we highlight that limiting the vocabulary of the language benefits the models with $\leq 22$ million parameters.

One possible explanation for this is the relative contribution of embedding parameters to model size. In an unconstrained setting, vocabulary embedding parameters account for most of the model, with no parameters left for transformer blocks.\footnote{For example, for a model with 10M parameters, if the embedding size is 200 with a vocabulary of 50,000, no transformer block can be added. But if the vocabulary size is 20,000, 6M parameters can be used in transformer blocks.
%In models with $\leq 10$M parameters, embedding parameters account for more than 50\% of the model parameters
%take up a larger portion of the total model size, allowing minimal transformer blocks.
} There is prior work showing that pre-training loss improves only minimally for models with zero transformer blocks \cite{kaplan2020scaling}. Thus, constraining vocabulary allows one to increase transformer block capacity while otherwise maintaining a small parameter count.

\section{Conclusions \& Future Work}
% In conclusion, this study found that the advantages of pre-training are observable even for smaller models, starting with a 1.25M parameter model. Our results also indicate that models with fewer layers achieve better performance on GLUE tasks, and that there is a strong correlation between upstream and downstream performance. However, we also found that pre-training compute optimality does not appear to be crucial for downstream results and that parameter count does not reliably predict upstream performance. Furthermore, we observed a break of the FLOP-Perplexity law at $2 \times 10^{15}$ FLOP region, which contradicts prior research.
% Overall, this study provides insights into the effects of downscaling large language models and highlights the importance of considering model shape and configuration when pre-training at a smaller scale. Our findings suggest that pre-training on a reduced vocabulary, similar to the way humans acquire language, may be an effective approach for training smaller models. Further research is needed to better understand the factors that affect the emergence of model abilities at a smaller scale and the implications of these findings for real-world applications.
In this study, we investigated whether reducing language size allows the benefits of pre-training to be observed in a downscaled setting for models with $\leq 20$M parameters.
%how reducing language size affects pre-training and downstream performance, for models with $\leq 20$M parameters. 
We evaluated a range of model configurations 
%and architectures 
%on a diverse set of datasets 
and found that the advantages of pre-training are observable even for models with as few as 1.25M parameters, with a strong correlation between upstream and downstream performance. However, we also observed that compute-optimal training does not appear to be crucial for downstream results and that parameter count does not reliably predict upstream performance. Furthermore, we observed a break of the FLOP-Perplexity power law at the $2.2 \times 10^{15}$ FLOP region, which shows the limited applicability of scaling laws.

Overall, our experiments provide insight into the behavior of small language models in a downscaled language setting. 
%highlighting the importance of considering model shape and configuration. 
The next logical steps as a follow-on to this work would be to check whether generative models would demonstrate any emergent abilities in a downscaled setting.

%Future work aimed at exploring small language model behavior should assess whether generative models similarly demonstrate emergent abilities at a smaller scale. 

% The next logical steps as a follow-on to this work would be to check whether generative models would demonstrate any emergent abilities in the downscaled setting.

% Additionally, our findings have important implications for real-world applications as they suggest that large models may not always be necessary for optimal performance.
% Future research should aim to further investigate these observations and to better understand the factors that affect the emergence of model abilities at a smaller scale.

\section{Limitations}
While we do explore a range of models in the 1-20M parameter space, our work does not constitute a complete study of downscaling. In this work, we aimed to explore the more fundamental components of model shape, model size, and input data. 

However, our findings may not generalize to other models with alternative applications of downscaling methods. Considering it to be out of scope for this study's assessment of pre-training effects, we did not compare our results to knowledge distillation methods of similar model shape and size. Furthermore, our exploration of model shape and size was limited to a model's hidden size, number of hidden layers, embedding size, and intermediate size, and number of attention heads as these are the most commonly-tuned hyperparameters.

Our usage of vocabulary filtration as a means of downscaling input data size may not be the most effective means of limiting input data. While shown to be effective, alternative approaches for input data manipulation such as curriculum learning, and data pruning merit study beyond the scope of this paper. 

\section*{Ethics Statement}
Our exploration of smaller language models presents a number of implications for accessibility, environmental impact, and cost. By exploring models in the space of 1-20M parameters, our findings can inform language modeling work for those without access to large, GPU-enabled environments. This is important as it can encourage further research work in this space by those who are otherwise unable to work with SoTA LLMs. We acknowledge that our resources enabled the breadth of study in this paper; most of this study was conducted using a single GPU. This consideration underscores our commitment to improving accessibility for under-resourced technologists throughout the world. Furthermore, in working with downscaled LLMs, we hope to encourage methods that reduce overall carbon footprint and bolster sustainable practices in NLP. These considerations are especially important given the particular burden placed on those with limited access to electricity and technology. The cost of running and experimenting with these models may prove quite costly in terms of person-hours and compute resources. As such, we hope our work at smaller scale can help lessen these burdens, and positively impact the lives of technologists, and others. Any model from the study can be trained in less than a day on a single consumer-grade GPU.

% \section*{Acknowledgements}

\bibliography{anthology,custom}
\bibliographystyle{acl_natbib}

\appendix

\section{Calculation of compute cost (FLOPs)}
\label{appendix:FLOPs-calculation}

We adopt the same approach for calculating the compute cost (FLOPs) as presented by \citet{hoffmann-optimal}. For notational convenience, we denote the sequence length, vocabulary size, embedding size, hidden size, intermediate size (hidden dimension of the feed-forward network in the transformer block), number of attention heads, key size (for the attention block), and number of layers by S, V, E, H, I, A, K and L, respectively.

The FLOPs for the forward pass are calculated as follows.

\begin{itemize}
    \item Single embedding block: 
        \begin{equation*}
            C_{emb} = 2 \times S \times (VE + EH)
        \end{equation*}

    \item Single attention block:\\
    
        - Cost of the key, query, and value projections
        \begin{equation*}
            C_{att} = 2 \times 3 \times SH \times (KA)
        \end{equation*}

        - Cost of the dot product operation of key and query
        \begin{equation*}
            C_{att} \mathrel{+}= 2 \times SS \times (KA)
        \end{equation*}

        - Cost of the softmax operation
        \begin{equation*}
            C_{att} \mathrel{+}= 3 \times SS \times (A)
        \end{equation*}

        - Cost of the query reduction
        \begin{equation*}
            C_{att} \mathrel{+}= 2 \times SS \times (KA)
        \end{equation*}

        - Cost of the final linear layer
        \begin{equation*}
            C_{att} \mathrel{+}= 2 \times SH \times (KA)
        \end{equation*}

    \item Single feed-forward layer:
        \begin{equation*}
            C_{int} = 2 \times (HI + IH)
        \end{equation*}

    \item Single language model head:
        \begin{equation*}
            C_{lmh} = 2 \times SHV
        \end{equation*}

    \item Single forward pass:
        \begin{equation*}
            C_{forward} = C_{emb} + C_{lmh} + L \times (C_{att} + C_{int})
        \end{equation*}

    \item Single backward pass:
        \begin{equation*}
            C_{backward} = 2 \times C_{forward}
        \end{equation*}

    \item Cost per training sequence:
        \begin{equation*}
            C_{seq} = C_{forward} + C_{backward}
        \end{equation*}
    
\end{itemize}

Therefore, we calculate the total compute cost as $Number\ of\ parameter\ updates\ \times batch\ size\ \times C_{seq}$.

\section{Tokenizer selection}
\label{appendix:tokenizer}

\subsection{List of reference words for ESMS}

In Table \ref{table:esms-word-list} we provide a list of words we used for calculating the Exact Sub-token Matching Score (ESMS). We also provide the morpheme-based tokens per word and the maximum value of exact matches per word.

\begin{table*}
\centering
\begin{tabular}{ccc}

\toprule

\textbf{Reference word} & \textbf{Morpheme sub-tokens} & \textbf{Maximum exact matches per word} \\

\midrule

{\centering Cooking} & cook, ing & 2  \\
Dangerous & danger, ous & 2  \\
Pretext & pre, text & 2  \\
Fitness & fit, ness & 2 \\
Antisocial & anti, social & 2 \\
Podium & pod, ium & 2 \\
Universe & uni, verse & 2 \\
European & europ, ean & 2 \\
Decode & de, code & 2 \\
Subvert & sub, vert & 2 \\
Proactive & pro, active & 2 \\
Concentric & con, centr, ic & 3 \\
Octopus & octo, pus & 2 \\

\bottomrule

\end{tabular}
\caption{Reference words and corresponding sub-word tokens for calculating Exact Sub-token Matching Score (ESMS) for the tokenized text}
\label{table:esms-word-list}

\end{table*}

\subsection{Comparison of different vocabulary sizes and tokenizer types}
We provide the results of our experiments to determine the best-suited tokenizer in this section. Table \ref{table:tokenizer-ESMS} provides the vocabulary size, corresponding word-split ratio, and ESMS value for the three types of tokenizers we evaluated. The bolded row is the final tokenizer we used in our pre-training experiments.

\begin{table*}
\centering
\begin{tabular}{clll}

\toprule 

\multicolumn{1}{p{6cm}}{\centering \textbf{Tokenizer name}}
& \multicolumn{1}{p{2cm}}{\centering \textbf{Vocabulary size}}
& \multicolumn{1}{p{2cm}}{\centering \textbf{Word-split ratio}}
& \multicolumn{1}{p{2cm}}{\centering \textbf{ESMS}} \\

\midrule

\multicolumn{1}{p{6cm}}{\centering BPE \cite{radford2019language}}
& \multicolumn{1}{p{2cm}}{\centering 18,000}
& \multicolumn{1}{p{2cm}}{\centering 1.34}
& \multicolumn{1}{p{2cm}}{\centering 0.2868} \\

\multicolumn{1}{p{6cm}}{\centering BPE \cite{radford2019language}}
& \multicolumn{1}{p{2cm}}{\centering \textbf{\underline{19,000}}}
& \multicolumn{1}{p{2cm}}{\centering \textbf{\underline{1.32}}}
& \multicolumn{1}{p{2cm}}{\centering \textbf{\underline{0.2604}}} \\

\multicolumn{1}{p{6cm}}{\centering BPE \cite{radford2019language}}
& \multicolumn{1}{p{2cm}}{\centering 20,000}
& \multicolumn{1}{p{2cm}}{\centering 1.31}
& \multicolumn{1}{p{2cm}}{\centering 0.2490} \\

\multicolumn{1}{p{6cm}}{\centering BPE \cite{radford2019language}}
& \multicolumn{1}{p{2cm}}{\centering Pre-trained}
& \multicolumn{1}{p{2cm}}{\centering 1.32}
& \multicolumn{1}{p{2cm}}{\centering 0.1547} \\

\midrule

\multicolumn{1}{p{6cm}}{\centering WordPiece \cite{devlin2018bert}}
& \multicolumn{1}{p{2cm}}{\centering \underline{16,000}}
& \multicolumn{1}{p{2cm}}{\centering \underline{1.17}}
& \multicolumn{1}{p{2cm}}{\centering \underline{0.0339}} \\

\multicolumn{1}{p{6cm}}{\centering WordPiece \cite{devlin2018bert}}
& \multicolumn{1}{p{2cm}}{\centering 17,000}
& \multicolumn{1}{p{2cm}}{\centering 1.17}
& \multicolumn{1}{p{2cm}}{\centering 0.0264} \\

\multicolumn{1}{p{6cm}}{\centering WordPiece \cite{devlin2018bert}}
& \multicolumn{1}{p{2cm}}{\centering 18,000}
& \multicolumn{1}{p{2cm}}{\centering 1.16}
& \multicolumn{1}{p{2cm}}{\centering 0.0188} \\

\multicolumn{1}{p{6cm}}{\centering WordPiece \cite{devlin2018bert}}
& \multicolumn{1}{p{2cm}}{\centering Pre-trained}
& \multicolumn{1}{p{2cm}}{\centering 1.17}
& \multicolumn{1}{p{2cm}}{\centering 0.0339} \\

\midrule

\multicolumn{1}{p{6cm}}{\centering SentencePiece \cite{raffel2020exploring}}
& \multicolumn{1}{p{2cm}}{\centering 9,000}
& \multicolumn{1}{p{2cm}}{\centering 1.32}
& \multicolumn{1}{p{2cm}}{\centering 0.0301} \\

\multicolumn{1}{p{6cm}}{\centering SentencePiece \cite{raffel2020exploring}}
& \multicolumn{1}{p{2cm}}{\centering \underline{10,000}}
& \multicolumn{1}{p{2cm}}{\centering \underline{1.29}}
& \multicolumn{1}{p{2cm}}{\centering \underline{0.0226}} \\

\multicolumn{1}{p{6cm}}{\centering SentencePiece \cite{raffel2020exploring}}
& \multicolumn{1}{p{2cm}}{\centering 11,000}
& \multicolumn{1}{p{2cm}}{\centering 1.26}
& \multicolumn{1}{p{2cm}}{\centering 0.0188} \\

\multicolumn{1}{p{6cm}}{\centering SentencePiece \cite{raffel2020exploring}}
& \multicolumn{1}{p{2cm}}{\centering Pre-trained}
& \multicolumn{1}{p{2cm}}{\centering 1.29}
& \multicolumn{1}{p{2cm}}{\centering 0.0339} \\

\bottomrule

\end{tabular}

\caption{Values of word-split ratio and ESMS for various tokenizer and vocabulary size settings.}
\label{table:tokenizer-ESMS}
\end{table*}

\section{Comparison with unconstrained language}
\label{appendix:comparison-with-unconstrained}

Our main experiments were conducted on language that is constrained by a predefined vocabulary. To study the effect of the applied vocabulary constraint in comparison with free text, we conduct a set of experiments on unconstrained language i.e., without any vocabulary-based filtering. In the following subsections, we provide details of the data collection, tokenizer training, and pre-training process adopted.

\subsection{Pre-training data}
Our objective in curating constrained language was solely to impose a vocabulary constraint. However, our filtering method (Section 3.2) resulted in constrained language comprised of non-consecutive text sequences. To address differences beyond vocabulary, we conducted unconstrained language collection using the following approach. We divided all instances in a specific corpus into spans of 110 words and randomly sampled spans. The number of randomly sampled spans was determined to maintain the same data distribution across different corpora, as indicated in Table \ref{table:data-collection}. This method aimed to minimize the impact of data features other than vocabulary. We gathered an equivalent number of training sequences (approximately nine million) as in the constrained pre-training data. Finally, we ensured a fair comparison of pre-training performance by using the same evaluation and test split for both pre-training datasets.

\subsection{Tokenizer}
After data curation, we conduct experiments with various tokenizers to finalize the tokenizer type and size of the token vocabulary for the language model. These experiments were conducted in the same manner described in Section 3.3. The final tokenizer we select is the Byte-BPE tokenizer \cite{radford2019language} with a vocabulary of 29,000 tokens ($1.6 \times$ that of the vocabulary size for the constrained language). The word-split ratio and the ESMS (exact sub-token matching score) for the final tokenizer were $1.53$ and $0.2339$.

\subsection{Experimental setup}
After finalizing the token vocabulary, we measure the pre-training as well as the downstream performance of the models trained on unconstrained language. We focus on the model configuration explored in the \textbf{set-1} (refer to section 4.1). Furthermore, guided by our results in the ICER analysis (refer to section 5.2), we only consider the model configurations that either perturb the hidden size or the number of layers in the model. With such a selection, we pre-train seven language models. In addition, we pre-train the smallest model configuration that highlighted the benefits of pre-training in our main experiment (refer to Table \ref{table:result-GLUE}). Overall, we pre-train 10 models on the collected unconstrained language. We keep all the hyperparameter values the same as in our main experiments (refer to Sections \ref{pt-ex} and \ref{ft-ex}). For the comparison of the pre-training performance, we measure the MLM loss and perplexity values calculated on the test split of the limited-vocabulary pre-training data. For comparison of the downstream performance, we finetune the final checkpoint of all pre-trained models on GLUE tasks and record the average GLUE score, separately for filtered and unfiltered versions of the GLUE datasets.

\section{Training larger models}
\label{appendix:larger-models}
We continued our pre-training experiments with the constrained language (limited vocabulary) data to include larger models i.e., models with more than 20 million parameters. We first set the anchor configuration to have embedding size, hidden size, intermediate size, number of layers, and number of attention heads equal to 512, 512, 2048, 8, and 8, respectively. After defining the anchor configuration, we follow the same approach of varying each configuration feature to explore and pre-train various models. However, for the training of larger models, we only focused on the hidden size and number of layers.

We present the pre-training results of the larger models in Table \ref{table:appendix-larger-models}. In the larger model configurations, we observe that the additional model parameters do not reduce the model perplexity below 4.67. However, within the set of larger models, we observe an expected reduction in the perplexity values with an increase in the size of the model. 

For our main experiments, we calculated the training data size for the expected largest model size i.e., 20 million parameters based on the findings provided by \citet{hoffmann-optimal}. The data size value was between 500 to 600 million tokens. Note that this data size is the size required to train the 20 million parameter model ‘compute-optimally’. Hence, we collected more data to observe the effect after the compute-optimal point. Finally, we collected approximately double the quantity of data ($\geq$ 1100 million tokens). Findings provided by \cite{hoffmann-optimal} are based on decoder-only models but, at the time of the experimentation, this was the best available guide for us to make a decision. Hence, we speculate that the size of our filtered pre-training data is not sufficient for the larger models that we consider in this set of experiments where we pre-train considerably larger models. Therefore, we do not include the model configuration considered in this set of experiments for our main results and figures.

\begin{table*}[th!]
\begin{small}
\centering
\begin{tabular}{cccc}

\toprule

  \textbf{\makecell{Model config. \\ (E, H, I, L, A)}}
& \textbf{\makecell{Model size \\ (mil. parameters)}}
& \textbf{\makecell{FLOPs \\ ($\times 10^{15}$)}}
& \textbf{\makecell{Perplexity}} \\
%& \textbf{\makecell{\% $\Delta$ GLUE score \\ (filtered)}}
%& \textbf{\makecell{\% $\Delta$ GLUE score \\ (unfiltered)}} \\

\midrule

(256, 256, 1024, 8, 2) & 16.24 & 110 & 4.67 \\

\midrule

(512, 512, 2048, 8, 8) & 45.30 & 302 & 5.57 \\

\midrule

(512, \textbf{256}, 2048, 8, 8) & 25.41 & 174 & 6.47 \\
(512, \textbf{768}, 2048, 8, 8) & 69.52 & 452 & 5.22 \\
(512, \textbf{1024}, 2048, 8, 8) & 98.06 & 624 & 4.94 \\

\midrule

(512, 512, 2048, \textbf{1}, 8) & 23.23 & 158 & 13.27 \\
(512, 512, 2048, \textbf{2}, 8) & 26.38 & 179 & 6.67 \\
(512, 512, 2048, \textbf{4}, 8) & 32.69 & 220 & 5.75 \\

\bottomrule

\end{tabular}

\caption{Pre-training results of larger models trained on the limited-vocabulary text data. In this table, we provide model configuration, model size, compute-cost (FLOPs), and model perplexity for the pre-training experiments we conducted on language models larger than the models pre-trained in our main experiments (set-1). We provide the best pre-training performance found in our main experiments (set-1) as a baseline (the first row). We find that additional parameters do not reduce model perplexity.}
\label{table:appendix-larger-models}
\end{small}
\end{table*}

\end{document}